\documentclass[10pt,twocolumn,letterpaper]{article}

\usepackage{cvpr}
\definecolor{cvprblue}{rgb}{0.21,0.49,0.74}
\usepackage[pagebackref,breaklinks,colorlinks,allcolors=cvprblue]{hyperref}

\def\paperID{*****} 
\def\confName{CVPR}
\def\confYear{2026}
 
\title{Diverse Yet Consistent: Context-Guided Diffusion with Energy-Based Joint Refinement for Multi-Agent Motion Prediction}

\author{Lei Chu\thanks{Equal contribution, this work was done when the authors were at USC. }, \ and Yuhuan Zhao\footnotemark[1]   \\
University of Southern California,
\\
Los Angeles, CA, USA\\
{\ttfamily\small \{lc\_285, yuhuanzh\}@usc.edu}
}

\begin{document}
\maketitle
\begin{abstract}
Deep generative models have become a promising approach for human motion prediction due to their ability to capture multimodal distributions and represent diverse human behaviors. However, generating predictions that are both diverse and jointly consistent among interacting agents remains challenging. In addition, most existing approaches are primarily evaluated using single-agent (marginal) metrics, which fail to fully reflect the joint dynamics of multi-agent interactions. We propose a diffusion-based framework that improves multi-agent motion prediction by leveraging rich contextual information from historical trajectories. This information is incorporated through a guidance mechanism to enhance the diversity and expressiveness of predicted motions. To further enforce interaction consistency, we introduce an energy-based formulation that refines the joint trajectory distribution while preserving the plausibility of individual trajectories.  Extensive experiments on four benchmark datasets demonstrate that our approach consistently outperforms existing methods. Notably, our approach substantially improves both marginal (ADE/FDE) and joint (JADE/JFDE) metrics on ETH/UCY over strong marginal baselines. Compared with prior joint prediction methods, it delivers significant gains in marginal metrics while maintaining competitive joint performance.
\end{abstract}    
\section{Introduction}
\label{sec:intro}

Human trajectory forecasting aims to predict future human movements while accounting for the uncertainty and diversity of possible behaviors \cite{alahi2016social, Deo2018C, Thiede2019A, salzmann2020trajectron++, mangalam2021goals, meng2022forecasting}. It plays a critical role in applications such as autonomous driving \cite{song2020pip, salzmann2020trajectron++, Cui2021L,  liu2023self, CHU2026Con}, digital health \cite{kothari2021human, mangalam2021goals}, and human–robot interaction \cite{sampieri2022pose, xia2024timestamp, zhang2024dynamic,  xia2025envposer}, where accurately anticipating pedestrian motion is essential for safe decision-making. In multi-agent environments \cite{salzmann2020trajectron++, jeong2024multi}, this task requires predicting a distribution of possible future trajectories by modeling both individual motion histories and interactions among agents \cite{zhao2019multi}. Despite significant progress, the problem remains challenging due to the inherent multi modality of human motion, where the same observed past can correspond to multiple plausible future paths across different environments \cite{jeong2025multi}.

Classical generative models, such as autoencoder-based approaches, address this challenge by learning compact latent representations of past trajectories and using them to decode and predict future motion. These models efficiently capture motion patterns and can be trained in unsupervised or self-supervised settings, making them easy to extend with techniques such as variational autoencoders \cite{walker2016uncertain, xu2022socialvae}, attention mechanisms \cite{huang2019stgat, Wong2020V}, or social interaction modules \cite{alahi2016social, mohamed2020social, xu2022socialvae}. However, they often have difficulty capturing multimodal future possibilities, which can lead to averaged trajectories and accumulated errors in long-term predictions. In addition, modeling complex multi-agent interactions often requires extra architectural components. In contrast, diffusion-based generative models \cite{bae2024singulartrajectory, tanke2023social, li2023bcdiff, mao2023leapfrog, jiang2023motiondiffuser, fu2025moflow} naturally model multimodal trajectory distributions by generating multiple plausible futures instead of a single averaged prediction. They can better capture complex motion patterns and interactions, producing diverse and realistic trajectories in crowded or uncertain environments. Nevertheless, diffusion models can exhibit temporal inconsistencies, where stochastic denoising introduces small jitters between time steps, and goal ambiguity, generating trajectories that appear locally plausible but are globally inconsistent with the underlying intent.

To overcome the challenges discussed above, we propose a deep generative modeling approach for human trajectory prediction. The main contributions of this paper are summarized as follows: 

\begin{itemize}
\item We propose \textbf{CODA}, a novel framework for consistent and diverse multi-agent motion prediction. Our approach enhances trajectory generation by incorporating rich contextual information from historical observations and integrating it into the generative process to improve prediction diversity and expressiveness.
\item We introduce a simple yet effective energy-based formulation \cite{du2019implicit} for joint trajectory refinement, which preserves the plausibility of individual trajectories while improving joint consistency among interacting agents.
\item Extensive experiments on four human motion datasets demonstrate that CODA achieves state-of-the-art performance across multiple metrics, effectively balancing the trade-off between marginal metrics (ADE/FDE) and joint metrics (JADE/JFDE) \cite{weng2023joint}, highlighting the importance of modeling diverse yet consistent multi-agent motion.
\end{itemize}

\begin{figure*}[ht]
  \centering
  \begin{subfigure}[t]{0.245\linewidth}
    \centering    \includegraphics[width=\linewidth]{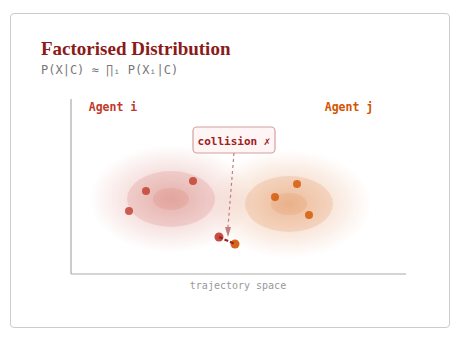}
    \caption{}
  \end{subfigure}
  \hfill
  \begin{subfigure}[t]{0.245\linewidth}
    \centering    \includegraphics[width=\linewidth]{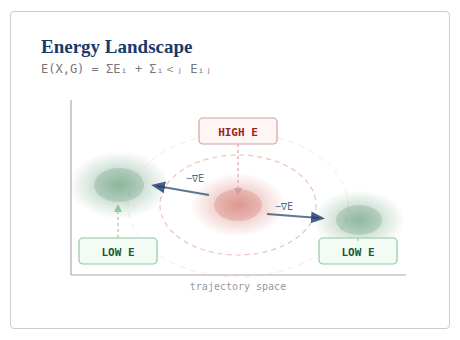}
    \caption{}
  \end{subfigure}
  \hfill
  \begin{subfigure}[t]{0.245\linewidth}
    \centering    \includegraphics[width=\linewidth]{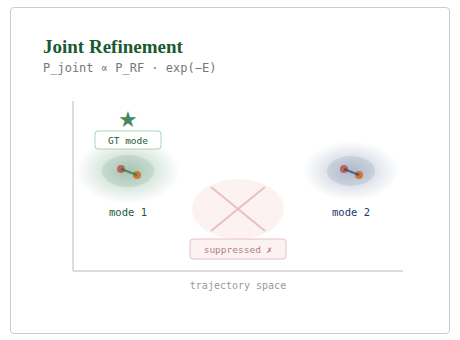}
    \caption{}
  \end{subfigure}
  \begin{subfigure}[t]{0.245\linewidth}
    \centering   \includegraphics[width=\linewidth]{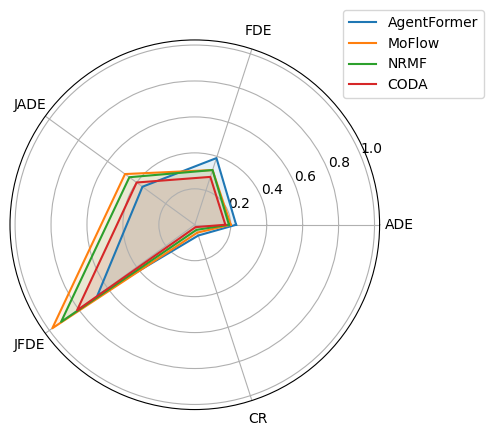}
    \caption{}
  \end{subfigure}
  \caption{Core concept and result of CODA: By incorporating rich interaction context and applying energy-based optimization, CODA improves joint behavior while preserving marginal accuracy, achieving the best performance on marginal metrics (ADE/FDE) and mean Collision Rate (CR), and the second-best results on joint metrics (JADE/JFDE).
  }
  \label{fig:core}
\end{figure*}
\section{Related Works}

\subsection{Human Trajectory Prediction} 

Human trajectory prediction aims to forecast future pedestrian positions from observed motion histories and surrounding interactions. Early approaches relied on physics-based and probabilistic models, such as the Social Force Model\cite{rudenko2018joint}, Kalman Filters \cite{meng2022forecasting}, and Hidden Markov Models \cite{qiao2014self, fang2025neuralized}, which describe motion using predefined dynamics but struggle to capture complex behaviors. With the advancement of deep learning, RNN- and LSTM-based models were introduced to learn temporal dependencies in trajectory sequences. To better model social interactions, subsequent work proposed interaction-aware architectures such as Social-LSTM \cite{alahi2016social} and graph-based methods using Graph Neural Networks (GNNs) \cite{mohamed2020social}. More recently, attention mechanisms and transformer-based models \cite{giuliari2021transformer, shi2023trajectory} have been explored to capture richer spatial–temporal dependencies and long-range interactions. In parallel, generative approaches, including GANs, VAEs, and diffusion models, have been developed to produce multiple plausible future trajectories, addressing the multi-modality of human motion prediction.

\subsection{Interaction Modeling and Enhancement}

Existing methods for interaction modeling in human trajectory prediction can be broadly categorized into several groups based on how agent interactions are represented. Rule-based approaches rely on hand-crafted formulations, such as the Social Force Model \cite{rudenko2018joint}, which describe pedestrian interactions using predefined physical rules \cite{huang2014action}. Neighborhood-based aggregation methods capture local interactions by pooling information from nearby agents, as exemplified by social pooling \cite{alahi2016social}. Pairwise interaction learning methods explicitly model relationships between agent pairs based on their relative positions and motion patterns \cite{yuan2020dlow,zhou2023query,janner2022planning}.

More recent work adopts graph-based methods, where Graph Neural Networks (GNNs) represent pedestrians as nodes and their interactions as edges \cite{wang2021graphtcn}, enabling flexible modeling of dynamic crowd structures. Finally, attention-based and transformer-based approaches further enhance interaction modeling by selectively focusing on relevant agents and capturing long-range spatial–temporal dependencies \cite{huang2019stgat,mohamed2020social,peng2021stirnet}. Building upon these developments, our work introduces richer contextual representations for diffusion-based trajectory generation, enabling diverse yet consistent multi-modal predictions.

\subsection{Generation with Guidance}

Guided generation in diffusion models is commonly implemented through classifier guidance, which adds gradients from an external classifier trained on noisy samples, and classifier-free guidance (CFG) \cite{ho2022classifier}, which combines conditional and unconditional score estimates without requiring a separate classifier. More generally, model- or regressor-guidance methods bias \cite{ho2020denoising, ho2022classifier, huberman2024edit} sampling using gradients from learned constraint models. In trajectory prediction, guidance is often realized through goal/intent guidance and scene/map guidance \cite{salimans2022progressive,lin2024progressive, adeli2021tripod}, which encourage trajectories to follow likely destinations while respecting environmental constraints \cite{rempe2023trace, yuan2023physdiff}. In this work, as illustrated in \cref{fig:core}, we adopt classifier-free guidance and incorporate richer contextual information during generation, while further refining the joint distribution of sampled trajectories with an energy-based model \cite{du2019implicit, ranzato2006efficient, pang2021trajectory}, preserving individual trajectory plausibility while improving multi-agent consistency \cite{weng2023joint}.

\section{Approach}

\subsection{Trajectory Prediction Problem}

Multi-agent trajectory forecasting aims to predict the future motion of multiple interacting agents based on their historical observations and scene context. Consider a dynamic scene containing $N$ agents observed over $T_h$ time steps. The historical trajectories of all agents are represented as
\begin{equation}
\mathbf{X}_{1:T_h} = \{\mathbf{x}_i^t \mid i = 1,\dots,N,\; t = 1,\dots,T_h \},
\end{equation}
where $\mathbf{x}_i^t \in \mathbb{R}^2$ denotes the 2D spatial position of agent $i$ at time step $t$. 
Given these observations, the objective is to predict the future trajectories of all agents over the next $T_f$ time steps,
\begin{equation}
\mathbf{Y}_{T_h+1:T_h+T_f} =
\{\mathbf{y}_i^t \mid t = T_h+1,\dots,T_h+T_f\},
\end{equation}
where $\mathbf{y}_i^t \in \mathbb{R}^2$ denotes the ground-truth future position of agent $i$ at time step $t$. The trajectory forecasting task can therefore be formulated as learning a model $f_\theta(\cdot)$ that estimates the conditional distribution
\begin{equation}
P_\theta(\mathbf{Y}_{T_h+1:T_h+T_f} \mid \mathbf{X}_{1:T_h}, \mathcal{C}),
\end{equation}
where $\mathcal{C}$ represents additional contextual information such as scene semantics, map priors, and agent interactions.
Since future trajectories are inherently uncertain and often multi-modal, the model typically predicts a set of $K$ possible future trajectories
$\hat{\mathbf{Y}}^{(k)}_{T_h+1:T_h+T_f}, \quad k = 1,\dots,K,$
each corresponding to a plausible future motion hypothesis.

\begin{figure*}[tb]
  \centering  \includegraphics[width=0.9\textwidth]{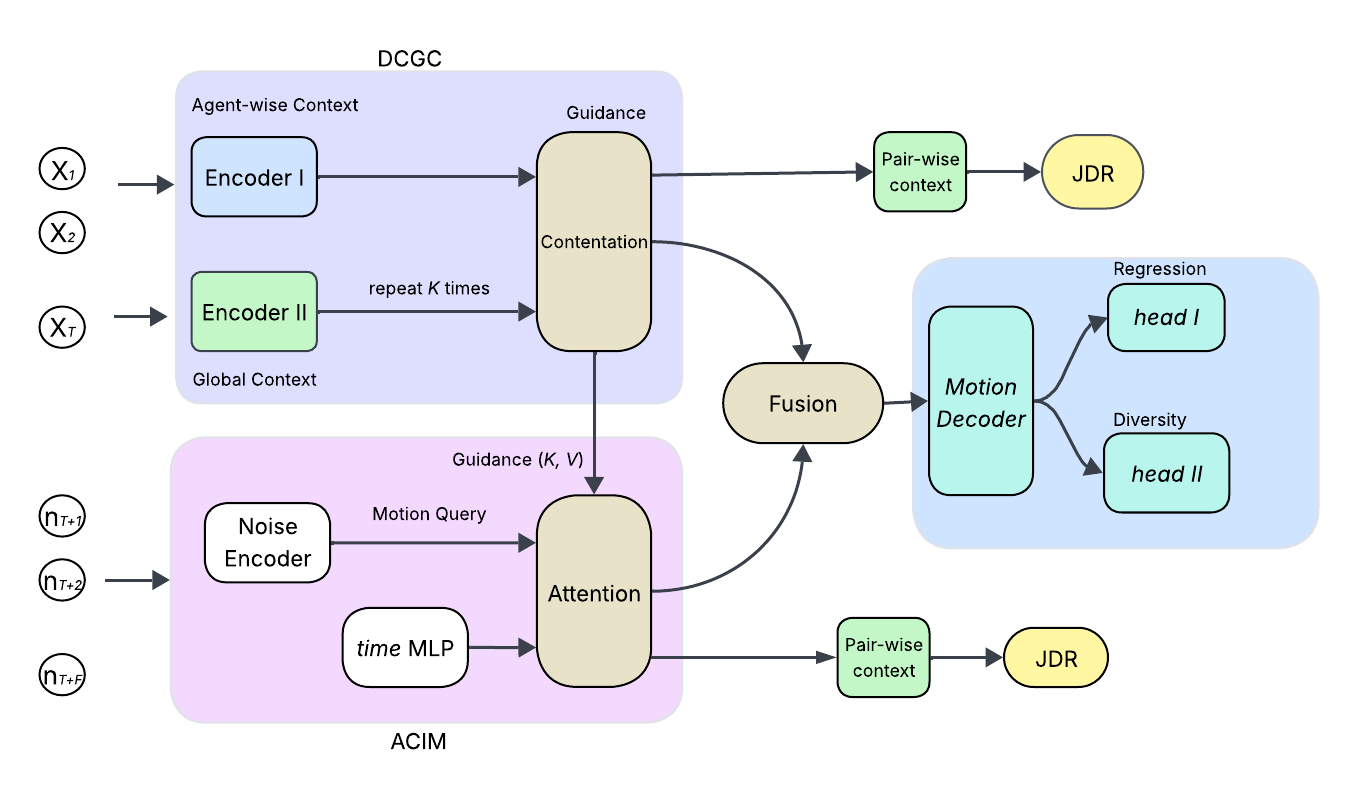}
  \caption{Illustrates the framework of our CODA. It consists of three key modules: (1) Dynamic Context as Guidance Condition (DCGC), which captures agents’ dynamic features as guidance conditions; (2) the Adaptive Condition Integration Module (ACIM), which incorporates agent dynamic context from these guidance conditions as additional noise to generate the next embedding during the diffusion generation phase; and (3) Joint Distribution Refinement (JDR), which shifts probability mass toward jointly consistent trajectories while reducing probability assigned to trajectories that are individually plausible but jointly inconsistent.
  }
  \label{fig:example}
\end{figure*}

In this section, we introduce the proposed method (\cref{fig:example}), CODA, a diffusion-based framework for joint trajectory modeling. CODA consists of three key modules: (1) Dynamic Context as Guidance Condition (DCGC), which extracts agents’ dynamic features as guidance conditions; (2) the Adaptive Condition Integration Module (ACIM), which injects the dynamic context as additional noise to guide embedding generation during diffusion; and (3) Joint Distribution Refinement (JDR), which shifts probability mass toward jointly consistent trajectories while suppressing trajectories that are individually plausible but jointly inconsistent.  

\subsection{Dynamic Context as Guidance Condition}

Recent studies construct guidance conditions from either agent interaction features (non-stationary) \cite{huang2019stgat, wang2021graphtcn} or dynamic features (stationary) \cite{huang2014action, adeli2021tripod, fang2025neuralized} to guide trajectory generation. However, both approaches have inherent limitations. Non-stationary conditions, represented by dynamic feature embeddings, improve trajectory diversity but often introduce redundant information that may cause deviations from the agent’s true intentions. In contrast, stationary conditions maintain intention consistency through agent interest embeddings, but typically rely on a fixed number of representations, which may fail to capture varying intention structures across agents and thus limit trajectory diversity. In this work, inspired by  \cite{yuan2021agentformer, xue2021mobtcast}, agent-wise features are obtained by applying self-attention \cite{lin2017structured} to embedded historical trajectories. Additionally, dynamic interaction features are extracted using the context Transformer \cite{xue2021mobtcast}.

\subsubsection{Agent-wise Context Extraction}
Given the embedded representation of the historical inputs $\mathbf{X}$, we employ a self-attention mechanism \cite{lin2017structured} to capture dependencies within the sequence. Specifically, a two-layer multilayer perceptron (MLP) is used to compute an attention score matrix that measures the relative importance of different historical observations. The weighted score matrix 
$\mathcal{W} \in \mathbb{R}^{A \times K}$ is computed as
\begin{equation}
\mathcal{W} =
\mathrm{Softmax}
\left(
\mathrm{MLP}_{4d \times K}
\left[
\tanh
\left(
\mathrm{MLP}_{d \times 4d}
\left(
\mathbf{X} + \mathcal{P}
\right)
\right)
\right]
\right),
\end{equation}
where $\mathcal{P}$ denotes the positional embedding and $k$ represents the number of extracted features, corresponding to the number of generated trajectory hypotheses.
The historical information is then aggregated using the weighted score matrix to obtain the interest feature representation:
\begin{equation}
\label{lcoal_con}
\mathbf{G}_{1}
\in \mathbb{R}^{k \times N \times d}
=
\mathcal{W}^{\top} \otimes \mathbf{X}.
\end{equation}

\subsubsection{Global Context Extraction}

The global context (dynamic interaction) embedding among agents is extracted using four standard Transformer blocks, similar to the architecture used in \cite{xue2021mobtcast}. Each block consists of a multi-head self-attention layer followed by a feed-forward network. The resulting global context representation is given by 
\begin{equation}
\label{global_con}
\mathbf{G}_{2}
\in \mathbb{R}^{1 \times N \times d}
=
\mathrm{Transformer}(\mathbf{X}),
\end{equation}
where $\mathbf{G}^{2}$ denotes the learned dynamic interaction embedding and $\mathrm{Transformer}(\cdot)$ represents the stacked Transformer blocks that model dependencies among agents. We repeat $\mathbf{G}^{2}$ for $K$ times for subsequent use.

Equations~\cref{lcoal_con} and~\cref{global_con} describe the extraction of the agent's inherent motion features and mutual interaction features, corresponding to approximations of long-term motion patterns and short-term interaction dynamics. Their combination captures the agent's true motion intention and guides the model to generate realistic trajectory embeddings in the noise space. To preserve the individuality of both features while enriching the guidance signal, we concatenate them to form the final guidance condition:
\begin{equation}
\label{global_all}
\mathbf{G} \in \mathbb{R}^{K \times A \times 2d} = Concat(\mathbf{G}_1,\mathbf{G}_2, dim=0),
\end{equation}

\subsection{Adaptive Context Integration}

Existing diffusion-based methods typically employ either MLPs or Transformer encoders for noise (or target) prediction. MLP-based approaches can improve trajectory diversity but lack explicit interaction with guidance conditions, which may introduce bias. In contrast, Transformer-based methods integrate noise representations with agent interaction features through attention weighting, effectively fitting noise in the agent intention space rather than the target space, thereby limiting trajectory diversity. 

To address this limitation, we propose an \textit{Adaptive Condition Integration Module} (ACIM) that dynamically injects the guidance condition $\bf G$ into the noisy target embedding via cross-attention, thereby strengthening conditional guidance during the diffusion process:
\begin{equation}
x_t^{G} =
\mathrm{Softmax}\!\left(
\frac{[(x_t + t_E + A_E)W^Q (GW^K)^{\top}]}{\sqrt{d}}
\right)
(GW^V),
\end{equation}
where $t_E$ and $A_E$ denote the time-step embedding and agent-order embedding, respectively. 
The guidance-enhanced representation is then fused with the original noisy embedding as
\begin{equation}
x_t = \mathrm{Concat}\left(x_t, x_t^{G}, \mathrm{dim}=-1\right).
\end{equation}
 
\subsection{Optimization with Joint Distribution Refinement}

\subsubsection{Context Guidance based Prediction}
Given contextual information $\mathbf{G}$, the goal is to estimate the conditional marginal distribution of the target variable $\mathbf{X}$, denoted as $p(\mathbf{X}\mid\mathbf{G})$. We adopt a conditional diffusion model \cite{ho2022classifier, rempe2023trace} that learns this distribution through a gradual denoising process. Specifically, a forward diffusion process progressively perturbs the data by adding Gaussian noise:

\begin{equation}
q(\mathbf{x}_t \mid \mathbf{x}_{t-1}) =
\mathcal{N}\!\left(
\sqrt{1-\beta_t}\mathbf{x}_{t-1}, \beta_t \mathbf{I}
\right), \quad t=1,\dots,T,
\end{equation}
which transforms the data distribution into an isotropic Gaussian. The reverse process learns to recover the clean sample conditioned on context $\mathbf{G}$:
\begin{equation}
p_\theta(\mathbf{x}_{t-1} \mid \mathbf{x}_t, \mathbf{G}),
\end{equation}
parameterized by a neural network that predicts the noise component $\epsilon_\theta(\mathbf{x}_t, t, \mathbf{G})$. The model is trained by minimizing the denoising objective
\begin{equation}
\label{noise_est}
\mathcal{L} =
\mathbb{E}_{t,\mathbf{x}_0,\epsilon}
\left[
\|\epsilon - \epsilon_\theta(\mathbf{x}_t, t, \mathbf{G})\|^2
\right],
\end{equation}
where $\epsilon \sim \mathcal{N}(0,\mathbf{I})$.
Through iterative denoising conditioned on $\mathbf{G}$, the model learns to sample from the target distribution $p(\mathbf{X}\mid\mathbf{G})$.

To follow the trajectory prediction formulation commonly used in the literature \cite{lin2024progressive, bae2024singulartrajectory, fu2025moflow, jeong2025multi}, we convert the noise prediction into target prediction and incorporate a best-of-K estimation along with a diversity-encouraging loss:
\begin{equation}
\label{loss_reg}
\begin{aligned}
L_{\mathrm{reg}} =\;&
\left\| \mathbf{Y} - \hat{\mathbf{Y}}(t,\mathbf{G}) \right\|_2^2
+ \min_{k \in \{1,\ldots,K\}}
\left\| \mathbf{Y} - \widehat{\mathbf{Y}}^{(k)} \right\|_2^2 \\
&+ \mathcal{L}_{\mathrm{div}}(\hat{\mathbf{Y}}, \mathbf{Y})
\end{aligned}
\end{equation}
In \cref{loss_reg}, the first term can be interpreted as a variant of  \cref{noise_est} \cite{lin2024progressive, bae2024singulartrajectory, fu2025moflow, jeong2025multi}. The second term corresponds to the well-known best-of-
$K$ estimation, while the final term represents the diversity loss. To encourage diversity among the $K$ predicted trajectory modes, we consider the temporal differences between consecutive predictions:$\Delta \hat{y}_i = \hat{y}_{i+1} - \hat{y}_i$ and $\Delta y_i = y_{i+1} - y_i. $
The diversity loss is defined as the Kullback-Leibler (KL) divergence between the normalized 
differences of predicted and ground-truth trajectories:
\begin{align}
\mathcal{L}_{\mathrm{div}}(\hat{\mathbf{Y}}, \mathbf{Y})
&=
D_{\mathrm{KL}}\!\left(
\sigma(\Delta \hat{\mathbf{y}})\; \|\; \sigma(\Delta \mathbf{y})
\right) \\
&=
\sum_{i=1}^{T_f-1}
\sigma(\Delta \hat{\mathbf{y}}_i)
\log
\frac{\sigma(\Delta \hat{\mathbf{y}}_i)}
{\sigma(\Delta \mathbf{y}_i)} .
\end{align}
The normalization function $\sigma(\cdot)$ is defined using a temperature-scaled
softmax:
\begin{align}
\sigma(\Delta y)_{i,j,k,l} &=
\frac{
\exp(\Delta y_{i,j,k,l}/\tau)
}{
\sum_{j'=1}^{K}
\sum_{k'=1}^{N}
\sum_{l'=1}^{T_f}
\exp(\Delta y_{i,j',k',l'}/\tau)
}, \\
\sigma(\Delta \hat{y})_{i,j,k,l} &=
\frac{
\exp(\Delta \hat{y}_{i,j,k,l}/\tau)
}{
\sum_{j'=1}^{K}
\sum_{k'=1}^{N}
\sum_{l'=1}^{T_f}
\exp(\Delta \hat{y}_{i,j',k',l'}/\tau)
}.
\end{align}

\subsubsection{Joint Distribution Refinement}

When the diffusion parameterization factorizes across agents, training yields accurate marginal distributions for each agent but does not explicitly capture their joint dependencies. This limitation motivates us to advocate joint distribution refinement (JDR).

In this work, we build on the observation that diffusion models define a distribution over trajectories and can learn accurate marginals. To address the missing joint dependencies, we introduce an energy-based model (EBM) \cite{du2019implicit}, parameterized by $\theta$, that provides a joint correction to the diffusion distribution. The resulting composed distribution is
\begin{equation}
\label{eq:refine}
P_{\text{joint}}(\mathbf{Y}  \mid \mathbf{G}) \propto 
\underbrace{P_{\text{Diff}}(\mathbf{Y} \mid \mathbf{G})}_{\text{good marginals}}
\underbrace{\exp\!\left(-E_{\theta}(\mathbf{Y}, \mathbf{G})\right)}_{\text{joint refinement}},
\end{equation}
where $P_{\text{Diff}}(\mathbf{Y} \mid \mathbf{G})$ provides \textbf{accurate marginal distributions} via  the diffusion model, whereas the energy term $\exp(-E_{\theta}(\mathbf{Y}, \mathbf{G}))$ acts as an EBM-based correction to better capture  \textbf{inter-agent dependencies}. 
Taking the logarithm of \cref{eq:refine} and focusing on the joint refinement part, we obtain
\begin{equation}
\label{eq:ebm}
{{\cal L}_{{\rm{EBM}}}} = \mathop {\min }\limits_\theta  \sum\limits_i {{E_\theta }} ({{\bf{Y}}_i},{{\bf{G}}_i}) + \sum\limits_{i < j} {{E_\theta }} ({{\bf{Y}}_i},{{\bf{Y}}_j},{{\bf{G}}_{ij}}),
\end{equation}
where the first term evaluates the \textit{individual plausibility} of each agent's prediction by measuring its consistency with the learned marginal distribution. The second term, on the other hand, enforces \textit{joint consistency} among agents by leveraging the \textbf{pairwise interaction context}, ${{\bf{G}}_{ij}}$, which encodes relational dependencies between agents. This interaction-aware mechanism encourages the predicted trajectories to remain not only individually realistic but also mutually compatible within the multi-agent environment  \cite{du2019implicit, ranzato2006efficient, pang2021trajectory}. 
We note that the refinement in~\cref{eq:refine} is related to prior work \cite{dhariwal2021diffusion}, where diffusion models are improved using score-based gradients. In contrast, our method performs refinement at the distribution level rather than the score level.

With the analysis above, we obtain the training loss used for our network. The overall training objective can be expressed as:
\begin{equation}
\label{total_loss}
\mathcal{L} =
\lambda_{\text{reg}} \mathcal{L}_{\text{reg}} +
\lambda_{\text{div}} \mathcal{L}_{\text{div}} +
\lambda_{\text{EBM}} \mathcal{L}_{\text{EBM}} 
\end{equation}

\subsubsection{Model Optimization and Discussion}


In our approach, trajectories are first normalized using min–max scaling to map future relative motions to the range $[-1,1]$, which helps stabilize training dynamics. For most datasets, we adopt the Social Transformer as the backbone architecture, while an MLP backbone is used for the NBA dataset. The transformer-based encoders incorporate skip connections and share a common configuration with 128 hidden features, a feed-forward dimension of 512, eight attention heads, and four stacked layers. The diffusion model generates samples through a 100-step denoising ODE process. For the flow time scheduler, we employ a logit-normal distribution. All experiments are conducted on an NVIDIA GeForce RTX 5090 GPU using the AdamW optimizer in PyTorch, with a weight decay of 0.01.

From an efficiency perspective, the proposed formulation is compatible with existing one-step and few-step diffusion models. Empirically, we find that a student model implemented as a one-step diffusion model achieves performance comparable to that of the teacher model when trained using Maximum Likelihood Estimation (MLE)-based distillation objectives, including KL divergence, Maximum Mean Discrepancy (MMD), and Chamfer distance. As optimization efficiency is not the primary focus of this work, we omit additional case studies of the one-step variant.
\section{Experiments}
\label{sec:casestudies}

\subsection{Datasets}

We evaluate the proposed method on four widely used trajectory prediction benchmarks: ETH/UCY, SDD, NBA, and JRDB. The ETH/UCY dataset contains five subsets (ETH, HOTEL, UNIV, ZARA1, ZARA2); following the standard leave-one-out protocol, we train on four subsets and test on the remaining one, predicting 12 future frames (4.8 s) from 8 observed frames (3.2 s). The SDD dataset \cite{robicquet2016learning} provides bird’s-eye-view pedestrian trajectories in pixel coordinates without projection matrices; we predict 12 future frames from 8 observations and report results in both pixel and metric units. The NBA dataset \cite{cervone2014pointwise} contains trajectories of 10 players and the ball captured by the SportVU system; we predict 20 future frames (4.0 s) conditioned on 10 observed frames (2.0 s), where frequent abrupt intention changes make trajectories more complex than typical pedestrian scenarios. The JRDB dataset \cite{saadatnejad2023jrdb} is a large-scale egocentric benchmark collected by a mobile social robot across diverse environments; we use the Social-Transmotion split \cite{saadatnejad2023social} for deterministic evaluation and the official challenge splits for stochastic prediction. As trajectories are annotated in camera coordinates while the robot is moving, we convert them to a global frame using odometry from rosbags. Following the official protocol, the model predicts 12 future frames from 9 observations at 2.5 Hz. 

\subsection{Baselines}

We compare the proposed CODA model with several state-of-the-art approaches, including NPSN \cite{bae2022non}, S-GAN\cite{Gupta2018S}, GroupNet \cite{xu2022groupnet}, LED \cite{mao2023leapfrog}, TUTR \cite{shi2023trajectory}, EqMotion \cite{xu2023eqmotion}, EigenTraj \cite{bae2023eigentrajectory}, SingularTraj \cite{bae2024singulartrajectory}, Evo-Graph \cite{mohamed2020social}, Y-net \cite{mangalam2021goals}, PECNet \cite{Mangalam2020I}, SocialVAE \cite{xu2022socialvae}, MemoNet \cite{Xu2022R},  LRR \cite{Lin2024I}, MOFLOW \cite{fu2025moflow} and NMFT \cite{fang2025neuralized}, on datasets where comparable results are publicly available. For the most recent methods, MOFLOW \cite{fu2025moflow} and NMFT \cite{fang2025neuralized}, we reproduce the reported results using their official codebases to ensure a fair comparison. The performance of the remaining methods is taken directly from their respective publications. Note that the set of compared methods may vary across datasets depending on the availability of reported results.

\subsubsection{Evaluation protocols}

We employ both marginal (ADE and FDE) and joint (JADE and  JFDE) metrics for evaluation \cite{weng2023joint}, which differ in the order of aggregation over samples and agents. The marginal metrics compute the minimum over $K$ predicted trajectories of the average displacement across time steps and the final-step displacement, respectively, on a per-agent basis. In contrast, joint metrics first average displacement errors across all agents within each predicted sample and then select the minimum over $K$ samples. This reordering, though subtle, is critical, as it enforces sample-level consistency and prevents combining predictions of different agents from different samples during evaluation.

\subsection{Quantitative Results}

{\bf ETH/UCY}: \cref{tab:ade_fde_eth} reports the minimum ADE/FDE on ETH/UCY with $K=20$ samples. CODA achieves the best overall performance, obtaining the lowest average error (0.17/0.28), corresponding to a 15.0\% / 12.5\% improvement over MoFlow (0.20/0.32) and a 10.5\% / 12.5\% improvement over MRF (0.19/0.32). Our method achieves the best ADE on ETH (0.24) and UNIV (0.22), and the lowest or tied-lowest FDE on ZARA1 (0.25) and ZARA2 (0.22). Compared with earlier generative models such as S-GAN, PECNet, and MemoNet, the improvements are substantially larger (over 30–60\% on average), demonstrating significantly enhanced trajectory accuracy and long-term prediction fidelity. The consistent gains across scenes indicate stronger multimodal modeling and more effective sample selection under the marginal evaluation protocol. 

\cref{tab:jade_fde_eth} presents joint evaluation results, where the best sample is selected after averaging errors across all agents, imposing stricter multi-agent consistency. Under this setting, CODA remains competitive, achieving the best performance on UNIV (0.52 JADE) and strong results across other subsets, with a competitive overall average (0.40/0.81). While some methods (e.g., AgentFormer variants and Joint VV) perform well on specific scenes, our approach maintains stable performance across environments. Moreover, the relatively small gap between marginal and joint metrics suggests that CODA produces coherent multi-agent predictions within each sampled trajectory, rather than relying on per-agent best-case selection. Overall, CODA generates diverse and accurate trajectories while preserving strong inter-agent consistency, demonstrating robustness across evaluation protocols.

\begin{table}[t]
\centering
\tiny
\caption{
Minimum ADE/FDE comparison across datasets with $K$=20 samples. Lower values indicate better performance. The best results are highlighted in bold, and the second-best results are underlined.
}
\begin{tabular}{l cccccc}
\toprule
Methods 
& ETH  
& HOTEL 
& UNIV  
& ZARA1  
& ZARA2  
& Avg. \\
\midrule

S-GAN \cite{Gupta2018S}
& 0.88/1.66
& 0.46/0.92
& 0.64/1.34
& 0.38/0.82
& 0.29/0.60
& 0.53/1.07 \\
  
Trajectron++ \cite{salzmann2020trajectron++}
& 0.67/1.18
& 0.19/0.28
& 0.30/0.54
& 0.25/0.41
& 0.18/0.32
& 0.32/0.55 \\

PECNet \cite{Mangalam2020I}
& 0.56/0.99
& 0.19/0.33
& 0.34/0.63
& 0.24/0.47
& 0.18/0.35
& 0.30/0.55 \\

Y-Net \cite{mangalam2021goals}
&  0.40 / 0.57 
& 0.12/0.19
& 0.31/0.60
& 0.26/0.49
& 0.20/0.39
& 0.26/0.45 \\

MemoNet \cite{Xu2022R}
& 0.41/0.64
& \underline{0.11}/\underline{0.17}
& \underline{0.24}/ 0.43
& 0.18/0.32
& \underline{0.14}/0.25
&  0.22/0.36 \\

View Vertically \cite{Wong2020V}
& 0.57/0.69
& 0.12/0.19
& 0.29/0.50
& 0.20/0.36
& 0.15/0.26
& 0.27/0.40 \\

Joint View Vertically \cite{weng2023joint}
& 0.70/0.79
& 0.13/0.20
& 0.27/0.47
& 0.22/0.36
& \underline{0.14}/0.25
& 0.29/0.41 \\

AgentFormer \cite{yuan2021agentformer}
& 0.45/0.75
& \underline{0.14}/0.23
& 0.25/0.45
&  0.18 / 0.30 
&  \underline{0.14}/0.24
&  0.23/0.39 \\

Joint AgentFormer \cite{weng2023joint}
& 0.47/0.79
& \underline{0.14}/0.21
& 0.29/0.51
& 0.19/0.32
& \underline{0.14}/0.24
& 0.25/0.41 \\

LRR \cite{Lin2024I}
& N/A & N/A & N/A
& N/A &N/A
& N/A \\

MoFlow \cite{fu2025moflow}
& 0.40/0.57
& \underline{0.11/0.17}
& 0.23/\textbf{0.39}
& \textbf{0.15}/\underline{0.26} 
& \textbf{0.12/0.22}
& 0.20/\underline{0.32} \\

NRMF \cite{fang2025neuralized}
& \underline{0.26}/\textbf{0.37}
& \underline{0.11}/\underline{0.17}
&  0.28/\underline{0.49}
&  0.18/0.30 
&  \underline{0.14}/0.25
& \underline{0.19/0.32} \\

CODA (Ours)
& \textbf{0.24/0.37}
& \textbf{0.10/0.15}
& \textbf{0.22/0.39}
& \textbf{0.15/0.25}
& \textbf{0.12/0.22}
& \textbf{0.17/0.28} \\
\bottomrule
\end{tabular}
\label{tab:ade_fde_eth}
\end{table}

\begin{table}[t]
\centering
\tiny
\caption{Quantitative comparison using min JADE$_{20}$/JFDE$_{20}$ $\downarrow$ with $K=20$ samples)}
\begin{tabular}{l c cccccc}
\hline
Method & 
ETH & HOTEL & UNIV & ZARA1 & ZARA2 & Avg. \\
\hline
S-GAN  \cite{Gupta2018S}
& 0.92/1.7 & 0.48/0.95 & 0.74/1.57 & 0.44/1.0 & 0.36/0.79 & 0.59/1.21 \\

Trajectron++  \cite{salzmann2020trajectron++} 
& 0.73/1.3 & 0.24/0.42 & 0.61/1.32 & 0.36/0.71 & 0.29/0.63 & 0.45/0.87  \\

PECNet \cite{Mangalam2020I}
& 0.62/1.1 & 0.29/0.59 & 0.67/1.42 & 0.41/0.90 & 0.37/0.84 & 0.47/0.97  \\

Y-Net  \cite{mangalam2021goals} 
& 0.50/0.78 & 0.21/0.39 & 0.70/1.56 & 0.49/1.04 & 0.49/1.10 & 0.48/0.97  \\

MemoNet \cite{Xu2022R} 
& 0.50/0.86 & 0.22/0.42 & 0.69/1.47 & 0.35/0.72 & 0.39/0.86 & 0.43/0.87  \\

View Vertically   \cite{Wong2020V} 
& 0.56/\textbf{0.78} & 0.20/0.33 & 0.65/1.31 & 0.33/0.65 & 0.30/0.60 & 0.41/0.73  \\

Joint View Vertically \cite{weng2023joint}
& 0.65/0.84 & \textbf{0.19}/\textbf{0.31} & \textbf{0.52}/\textbf{1.09}
& 0.33/0.63 & 0.27/0.55 & 0.39/\underline{0.68}  \\

AgentFormer \cite{weng2023joint}
& \textbf{0.48}/0.79 & 0.24/0.46 & 0.62/1.31 & \underline{0.29}/\underline{0.56}
& 0.30/0.62 & \underline{0.38}/0.75 \\

 Joint AgentFormer  \cite{weng2023joint}
& \underline{0.49/0.80} & \underline{0.19}/\underline{0.32} &  0.59/1.22
& \textbf{0.27}/\textbf{0.51} & \textbf{0.25}/\textbf{0.51}
& \textbf{0.36}/\textbf{0.67}  \\

LRR \cite{Lin2024I}
& 0.51/0.91 & \underline{0.19}/\underline{0.33} &  0.61/1.25
& 0.33/0.65 & 0.28/0.59
& 0.39/0.74  \\

MoFlow \cite{fu2025moflow}
&  0.71/1.36 & 0.34/0.66  &  0.59/1.12
&  0.41/0.93  & 0.38/0.87
&   0.48/0.98 \\

NRMF \cite{fang2025neuralized}
&  0.57/1.11 
&  0.27/0.50
& 0.64/1.31
&  0.43/0.94
& 0.34/0.74
&  0.45/0.92 \\

CODA (Ours)
& 0.56/1.07  & 0.23/0.42 & {\bf{0.52}}/\underline{1.10} 
& 0.39/0.82  &  0.30/0.66
& 0.40/0.81 \\
\hline
\end{tabular}
\label{tab:jade_fde_eth}
\end{table}

\begin{table}[ht]
\centering
\tiny
\caption{
Comparison with state-of-the-art methods on the NBA dataset.  
}
\begin{tabular}{c|c|cccccc}
\hline
  & Time & GroupNet & MID & LED & MOFLOW & NRMF & CODA \\
\hline
ADE & 1.0s & 0.26/0.34 & 0.28/0.37 & 0.21/0.28 & 0.18/0.25 & \textbf{0.16}/\textbf{0.24} & \underline{0.17}/\textbf{0.24} \\
 /FDE & 2.0s & 0.49/0.70 & 0.51/0.72 & 0.44/0.64 & \underline{0.34/0.48} & \underline{0.34}/0.50 & \textbf{0.33/0.45} \\
  & 3.0s & 0.73/1.02 & 0.71/0.98 & 0.69/0.95 & \underline{0.51/0.68} & 0.53/0.75 & \textbf{0.50}/\textbf{0.66} \\
  & 4.0s & 0.96/1.30 & 0.96/1.27 & 0.94/1.21 & \underline{0.70}/\underline{0.89} & 0.75/0.97 & \textbf{0.69}/\textbf{0.87} \\
\hline
JADE & 1.0s & N/A & N/A & N/A & 0.37/0.68 & \textbf{0.33}/\textbf{0.61} & \underline{0.36}/\underline{0.67} \\
 /JFDE & 2.0s & N/A & N/A & N/A & 0.81/1.62 & {\textbf{0.73/1.46}} & \underline{0.79/1.59}  \\
  & 3.0s & N/A & N/A & N/A & 1.26/2.50 & \textbf{1.15/2.21} & \underline{1.23}/\underline{2.47} \\
  & 4.0s & N/A & N/A & N/A & 1.69/3.31 & {\textbf{1.53/2.79}} & \underline{1.67}/\underline{3.28} \\
\hline
\end{tabular}
\label{res:nba}
\end{table}

{\bf NBA:} \cref{res:nba} presents the temporal evaluation on the NBA dataset under both marginal (ADE/FDE) and joint (JADE/JFDE) metrics. Across all prediction horizons (1.0s–4.0s), CODA demonstrates consistently strong performance. Under marginal metrics, CODA achieves the best results at longer horizons, obtaining the lowest ADE/FDE at 3.0s (0.50/0.66) and 4.0s (0.69/0.87), while remaining competitive at shorter horizons. Although NRMF attains the best performance at 1.0s (0.16/0.24), CODA maintains comparable accuracy and outperforms the other baselines as the prediction horizon increases. Overall, these ADE/FDE results indicate that CODA provides the most robust marginal trajectory prediction, particularly for longer-term forecasting.

For the joint metrics, NRMF achieves the best results across all prediction horizons. CODA ranks second in most cases. The widening gap at longer horizons indicates that NRMF remains a strong competitor to CODA, and that both methods are effective at modeling long-term dynamics and complex inter-agent interactions in NBA trajectories. Overall, the JADE/JFDE results show that both methods model multi-agent dependencies well, with NRMF having a slight advantage in joint forecasting accuracy.

{\bf SDD}: \cref{tab:SD} reports the quantitative comparison on the SDD dataset under both marginal (ADE/FDE) and joint (JADE/JFDE) metrics. CODA achieves the best marginal performance, obtaining the lowest ADE (6.87) and FDE (10.86), outperforming strong recent baselines such as MRF (7.10/11.11), MoFlow (7.50/11.96), and ET+HighGraph (7.81/11.09). These improvements are notable given the highly dynamic nature of NBA trajectories, which involve abrupt motion changes and complex player interactions.

Under joint metrics, AgentFormer  achieves the best JADE (9.56) and Y-Net the lowest JFDE (16.01), indicating stronger optimization for joint trajectory coherence. In contrast, recent generative approaches—including MoFlow, NRMF, and CODA—primarily optimize marginal trajectory accuracy, which can reduce joint consistency. Overall, CODA establishes state-of-the-art marginal performance on NBA while maintaining competitive joint results, demonstrating its effectiveness in modeling complex, interaction-rich dynamics.

\begin{table}[t]
\centering
\footnotesize
\caption{
Comparison with state-of-the-art methods on the SDD dataset.
}
\setlength{\tabcolsep}{4pt}
\renewcommand{\arraystretch}{1.15}
\begin{tabular}{l cccccc   }
\toprule
Methods  
& Venue 
& ADE 
& FDE 
& JADE
& JFDE  
  \\
\midrule
S-GAN 
&  CVPR'18
& 12.74 
& 22.65
& 13.76 
& 24.84
  \\  
Trajectron++
&  ECCV'20
& 10.18 
& 15.76
& 11.36 
& 18.21
 
  \\

PECNet
&  ECCV'20
&  9.34 
&  16.10
& 10.82 
& 19.48
 
  \\

Y-Net
&   ICCV'21
& 8.15
&  12.80 
& \underline{9.67}
& \bf{16.01}
 
  \\

MemoNet
& CVPR'22
&   7.97 
& 12.82
& 9.59 
& \underline{16.43}
 
 \\

View Vertically
&  ECCV'22
& 9.34 
& 14.67
& 10.75 
& 17.45
 
  \\

 Joint View Vertically 
&  CVPR'23
& 9.62 
& 15.07
&  10.92 
& 17.70
 
  \\

AgentFormer
&  CVPR'21
& 8.01
&  13.24
& \underline{9.67 } 
& 16.92
 
 \\
Joint AgentFormer    
&  CVPR'23
& 8.25 
& 13.74
& \bf{9.56} 
&  16.59 
  \\
 TUTR  
& ICCV'23 & 7.76  & 12.69 
& -
& -  \\
 
 ET+HighGraph 
& CVPR'24 & 7.81 &11.09
& -
&  - \\

 MoFlow  
&  CVPR'25 & 7.50 & 11.96
&  18.97 & 37.46 
  \\

 NRMF  
&  ICLR'25 & \underline{7.10}   &  \underline{11.11}
&  16.18 &  32.55
  \\

 CODA (Ours)  
& TBD &  \bf{6.87}  &  \bf{10.86}
&  13.96 &  26.31
  \\

\bottomrule
\end{tabular}

\label{tab:SD}
\end{table}

{\bf JRDB:} \cref{res:jrdb} reports time-horizon evaluation on JRDB using both marginal (ADE/FDE) and joint (JADE/JFDE) metrics. Across prediction intervals from 1.2s to 4.8s, CODA consistently achieves competitive or superior performance. Under marginal metrics, it attains the best results at 1.2s (0.04/0.05) and remains tied for the lowest errors at longer horizons, reaching 0.11/0.17 at 3.6s and 0.15/0.23 at 4.8s, indicating strong short-term precision and stable long-term forecasting. Under joint metrics, CODA demonstrates improved multi-agent consistency over MOFLOW and NRMF, particularly at longer horizons; at 4.8s, it achieves the lowest JADE/JFDE (0.34/0.64), outperforming MOFLOW (0.35/0.67) and NRMF (0.37/0.69). The widening gap at extended horizons suggests better preservation of global scene coherence. Overall, CODA maintains strong accuracy across horizons while achieving superior long-term joint consistency, highlighting its effectiveness in modeling dynamic, egocentric multi-agent environments.

\begin{table}[ht]
\centering
\scriptsize
\caption{
Comparison with state-of-the-art methods on the JRDB dataset.  
}
\begin{tabular}{c|c|c|c|c|c}
\hline
&Time & LED & MOFLOW & NRMF & CODA \\
\hline
ADE/FDE&1.2s & 0.05/0.07 & 0.04/0.06 & \textbf{0.04/0.05} & \textbf{0.04}/\textbf{0.05} \\
&2.4s & 0.09/0.14 & \textbf{0.07/0.11} & \underline{0.08}/\textbf{0.11} & \textbf{0.07/0.11} \\
&3.6s & 0.14/0.21 & \textbf{0.11/0.17} & \textbf{0.11/0.17} & \textbf{0.11}/\textbf{0.17} \\
&4.8s (Total) & 0.18/0.28 & \textbf{0.15/0.23} & \textbf{0.15/0.23} & \textbf{0.15}/\textbf{0.23} \\
\hline
JADE/JFDE&1.2s & N/A  & 0.10/0.15 & \textbf{0.09}/0.16 & \textbf{0.09}/\textbf{0.14} \\
&2.4s & N/A & 0.18/0.32 & 0.19/0.35 & \textbf{0.17/0.31} \\
&3.6s & N/A & 0.27/0.50 & 0.28/0.52 & \textbf{0.25}/\textbf{0.47} \\
&4.8s (Total) & N/A & 0.35/0.67 & 0.37/0.69 & \textbf{0.34}/\textbf{0.64} \\
\hline
\end{tabular}
\label{res:jrdb}
\end{table}

\subsection{Qualitative  Results}

This section presents a qualitative comparison of multi-modal trajectory predictions for Scene 2, Agent 8 from the NBA dataset. Each column corresponds to a different method (MOFLOW, NRMF, and CODA), visualizing the observed trajectories, ground-truth futures, sampled predictions, and the mean predicted trajectory. The reported standard deviation reflects the dispersion of sampled trajectories and provides an indication of predictive diversity.

\begin{figure}[ht]
  \centering
  \includegraphics[width=\linewidth]{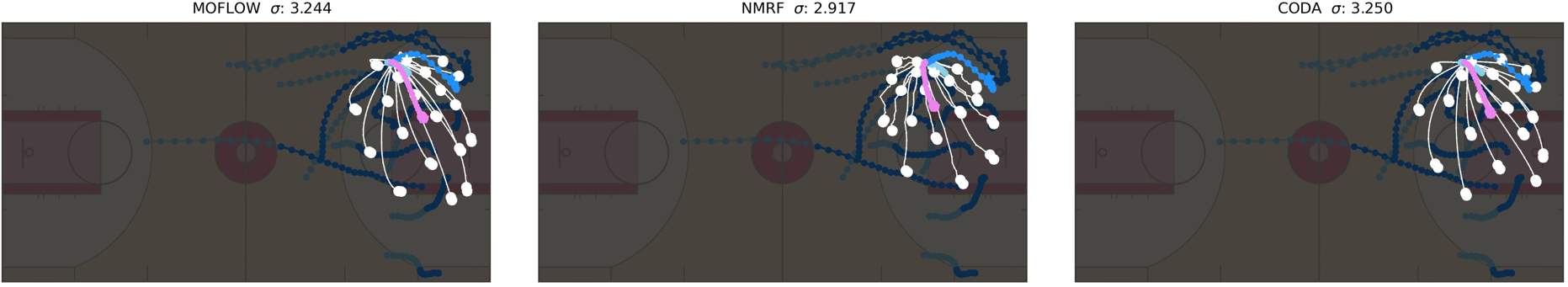}
  \caption{Prediction samples ($T_h$=20) on the NBA dataset. Light blue shows historical trajectories, dark blue shows future ground truth, white curves are sampled predictions, and the violet curve denotes the mean estimate.
  }
  \label{fig:qual_res}
\end{figure}

As shown in \cref{fig:qual_res}, MOFLOW generates diverse trajectories that largely align with the ground truth but exhibit mild dispersion in highly dynamic cases. NRMF produces more concentrated predictions with controlled variance while maintaining plausible future directions. CODA achieves similar diversity with better alignment to the ground truth and interaction context, resulting in improved structural coherence across agents. Overall, while all methods capture complex interaction-driven sports dynamics, CODA demonstrates stronger joint behavioral consistency.

\begin{figure}[ht]
  \centering
  \includegraphics[width=\linewidth]{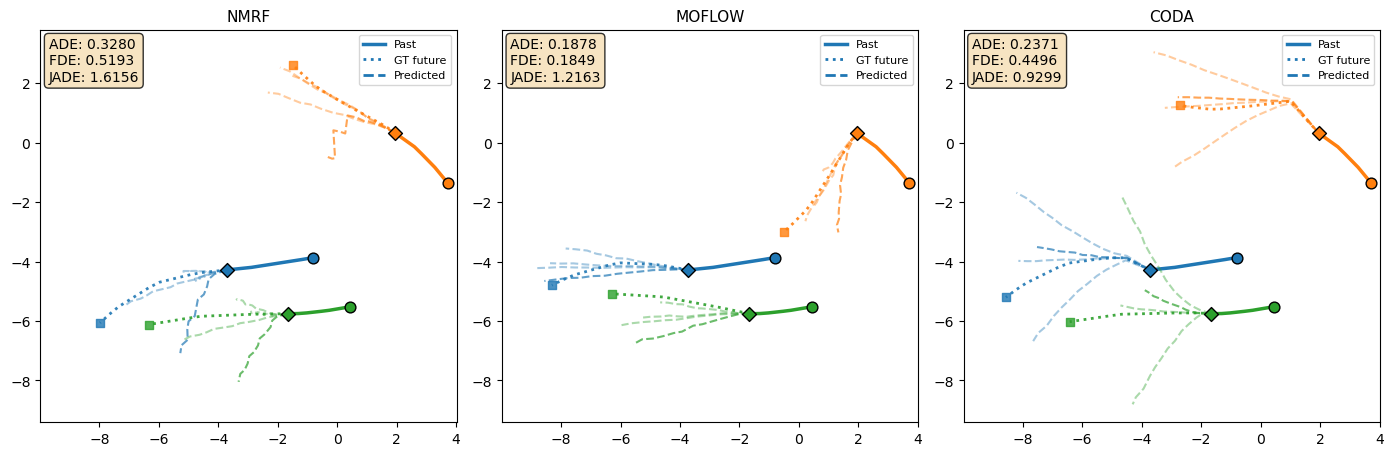}
  \caption{Qualitative trajectory prediction results on the Univ dataset. }
  \label{fig:qual_univ}
\end{figure}

Similar observations can be made in \cref{fig:qual_univ}. NRMF produces diverse trajectories but occasionally deviates from the dominant motion pattern, resulting in larger displacement errors. MOFLOW more closely follows the ground truth while preserving trajectory diversity. CODA strikes a balance between diversity and structural coherence, capturing the overall motion trend while maintaining interaction-consistent variations. Although its ADE (0.2371) is slightly higher than that of MOFLOW in this example, CODA achieves a lower JADE (0.9299), indicating improved joint behavioral consistency across agents.

\subsection{Ablation Study}

\begin{table}[t]
\centering
\scriptsize
\caption{Ablation study evaluating the components of the proposed method.}
\label{tab:ablation}
\begin{tabular}{c c c | c c}
\hline
DCGC  & ACIM  & JDR & ADE/FDE & JADE/JFDE \\
\hline
\checkmark & $\times$ & $\times$ & 0.740/0.923 & 1.809/3.485 \\
\checkmark & \checkmark & $\times$  & 0.724/0.901 & 1.764/3.421 \\
$\times$  & $\times$   &  \checkmark & 0.716/0.897 & 1.665/3.273 \\
\checkmark   & \checkmark & \checkmark & 0.694/0.873 & 1.671/3.283 \\
\hline
\end{tabular}
\end{table}

We conduct an ablation study on the NBA dataset to examine the contributions of the key components in our framework—DCGC, ACIM, and JDR—using both marginal metrics (ADE/FDE) and joint metrics (JADE/JFDE). As reported in \cref{tab:ablation}, removing DCGC or ACIM leads to clear increases in ADE and FDE, indicating that these modules play an essential role in improving agent-wise trajectory prediction by capturing richer contextual and interaction information. In contrast, removing JDR results in only minor changes in ADE/FDE but causes noticeable degradation in JADE/JFDE. This result suggests that JDR mainly contributes to modeling joint dynamics and coordination among agents. Overall, the ablation results demonstrate a clear functional distinction: DCGC and ACIM primarily enhance marginal prediction accuracy, whereas JDR is crucial for improving joint trajectory consistency.

\section{Conclusion}

In this work, we propose CODA, a framework for improving multi-agent motion prediction by incorporating rich contextual information from historical trajectories into the generative process, enabling more diverse and expressive predictions. We further introduce a simple yet effective formulation that refines the joint trajectory distribution using an energy-based model, preserving the plausibility of individual trajectories while enhancing joint consistency among interacting agents. Extensive experiments on four benchmark datasets demonstrate consistent improvements over state-of-the-art approaches. Overall, these results highlight the effectiveness of CODA in jointly modeling diversity and consistency for multi-agent motion prediction.

{\bf Limitations.} We leverage different levels of data-driven context, achieving SOTA performance on ADE/FDE while attaining competitive results on the joint metrics JADE, JFDE, and CR. However, the additional effort devoted to robust context modeling increases the training time.  All methods still have substantial room for improvement on the CR metric, although this issue could be significantly alleviated by incorporating LiDAR data. Furthermore, the current prediction pipeline relies solely on coordinate inputs; integrating richer physical context—such as traversable areas—could further enhance performance. 
{
    \small
    \bibliographystyle{ieeenat_fullname}
    \bibliography{main}
}


\end{document}